\documentclass{article}
\usepackage{spconf,amsmath,graphicx}
\usepackage[ruled,vlined]{algorithm2e}
\usepackage{amsmath,amsfonts,bm, bbm}

\newcommand{\vect}[1]{\mathbf{#1}}

\def\eqref#1{equation~\ref{#1}}
\def\1{\bm{1}}

\DeclareMathAlphabet{\mathsfit}{\encodingdefault}{\sfdefault}{m}{sl}
\SetMathAlphabet{\mathsfit}{bold}{\encodingdefault}{\sfdefault}{bx}{n}

\usepackage{mdframed}

\usepackage{accents}
\newlength{\dhatheight}

\usepackage{hyperref}
\usepackage{flushend}

\usepackage{booktabs}
\title{No More Than 6ft Apart: Robust K-means via Radius Upper Bounds}
\name{Ahmed Imtiaz Humayun$^{\dagger}$ \qquad Randall Balestriero$^{\ddagger}$ \qquad Anastasios Kyrillidis$^{\dagger}$ \qquad Richard Baraniuk$^{\dagger}$}{}
\address{$^{\dagger}$Rice University \qquad $^{\ddagger}$Meta AI Research}
\begin{document}
%\ninept
%
\maketitle
\begin{abstract}
Centroid based clustering methods such as k-means, k-medoids and k-centers are heavily applied as a go-to tool in exploratory data analysis. In many cases, those methods are used to obtain {\em representative} centroids of the data manifold for visualization or summarization of a dataset. Real world datasets often contain inherent abnormalities, e.g., repeated samples and sampling bias, that manifest imbalanced clustering. We propose to remedy such a scenario by introducing a maximal radius constraint $r$ on the clusters formed by the centroids, i.e., samples from the same cluster should not be more than $2r$ apart in terms of $\ell_2$ distance.
% This novel K-means formulation produces centroids with space distribution robust to dataset inconsistencies. 
We achieve this constraint by solving a semi-definite program, followed by a linear assignment problem with quadratic constraints.
Through qualitative results, we show that our proposed method is robust towards dataset imbalances and sampling artifacts. To the best of our knowledge, ours is the first constrained k-means clustering method with hard radius constraints.\footnote{\href{https://github.com/AhmedImtiazPrio/radius-constrained-kmeans}{\texttt{Codes at https://bit.ly/kmeans-constrained}}}
\end{abstract}
\begin{keywords}
robust k-means, radius constraint, constrained optimization, data imbalance, clustering
\end{keywords}

\section{Introduction}
\label{intro}

K-clustering methods offer the benefit of producing summarized dataset representations via a set of learned centroids or centers. Such representations find many applications from denoising, anomaly detection, visual summarization, as initial parameters for downstream algorithms such as Gaussian Mixture Models, and as plastic features for life-long machine learning classifiers \cite{hao2018lifelong}.
% All those downstream usage however rely on the ability of K-means to produce meaningful dataset representations.
The fundamental assumptions governing the success of K-means lie in having clusters with roughly the same number of samples and intra-cluster data covariance that is isotropic with the form $\sigma I$. Furthermore, $\sigma$ should be roughly the same between clusters. Whenever the data does not align with those assumptions, K-means gets skewed toward producing an incorrect representation. For example, even in the simplest case of having a dataset made of a mixture of Gaussians but with a varied number of samples per mixture, K-means centroids will naturally shift toward the mode with the greatest number of samples.

The implication of those cases can be dramatic as any downstream task relying on those representations, would be negatively impacted causing, e.g., bias in facial recognition models \cite{buolamwini2018gender}, gender bias in word-level language models \cite{bordia2019identifying}. This has led to the birth of many K-means alternatives, each aiming at fixing a particular limitation, e.g., the presence of outliers among others \cite{hesabi2015data, chiplunkar2020solve}.
There also exists k-clustering methods focused on (fair) data summarization \cite{kleindessner2019fair, bera2019fair, chierichetti2018fair}, imbalanced data clustering \cite{kumar2014imbalanced} and robustness to specific transformations of the data \cite{garcia1999robustness,honda2010fuzzy,dorabiala2021robust}. Some of these methods require specifications on the cardinality of the demographics \cite{kleindessner2019fair, bradley2000constrained, rujeerapaiboon2019size}, weak labels of imbalance \cite{kumar2014imbalanced}, the data transformations to be robust against \cite{dorabiala2021robust}, or a priori knowledge of the data/outlier distributions \cite{li2021tk}. 

\begin{figure}[t]
\centering
\includegraphics[width=0.85\linewidth]{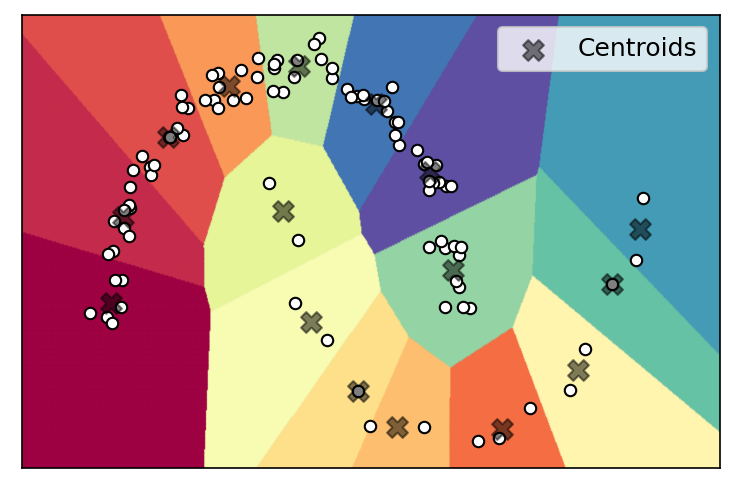}
\caption{\small Centroids generated by our proposed maximal radius constrained k-means method for $K=16$ on imbalanced two-moons data $(85/15)$. Even though the concave moon is oversampled more than ${5}$ times, our method produces equal number of centroids for both moons. See Fig~\ref{fig.tripletimbalmoons} for comparisons.
% with k-\{means,medoids,center\}.
% Moreover, the summaries have perceptually equal spacing along the data manifold. This indicates that the representants can possibly estimate a discretization of the data manifold while being robust towards sampling inconsistencies.
}
\label{fig:rconstimbalmoons}
\end{figure}

\begin{figure*}[t]
	\centering
    \begin{minipage}{.24\textwidth}
    \includegraphics[width=\linewidth]{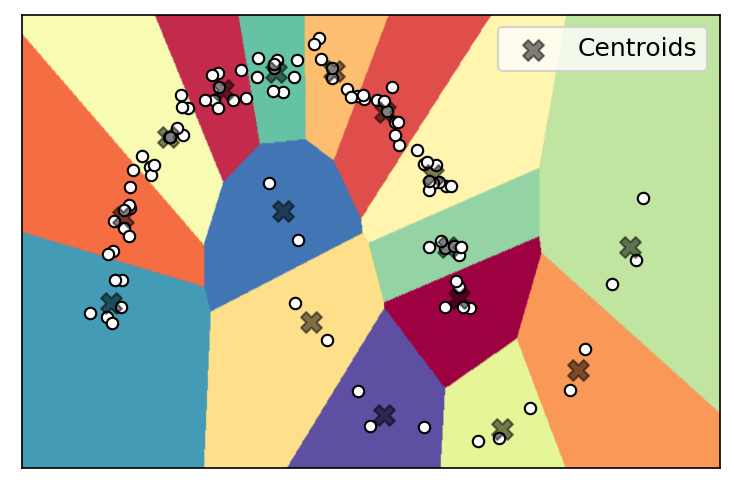}
    \end{minipage}
    \begin{minipage}{.24\textwidth}
    \includegraphics[width=\linewidth]{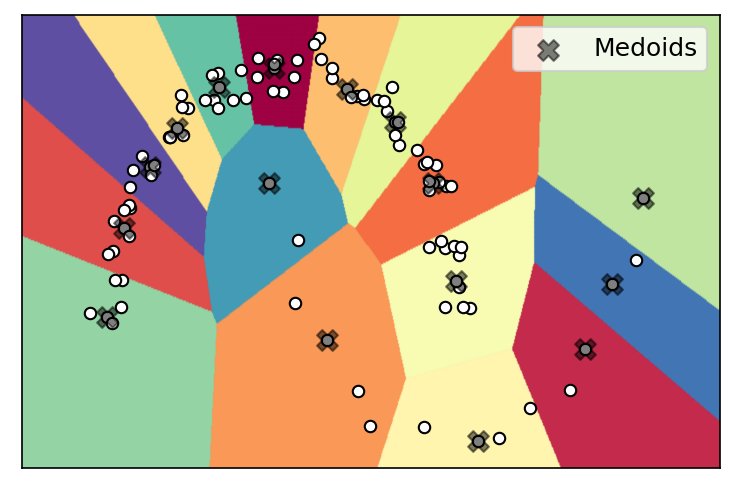}
    \end{minipage}
    \begin{minipage}{.24\textwidth}
    \includegraphics[width=\linewidth]{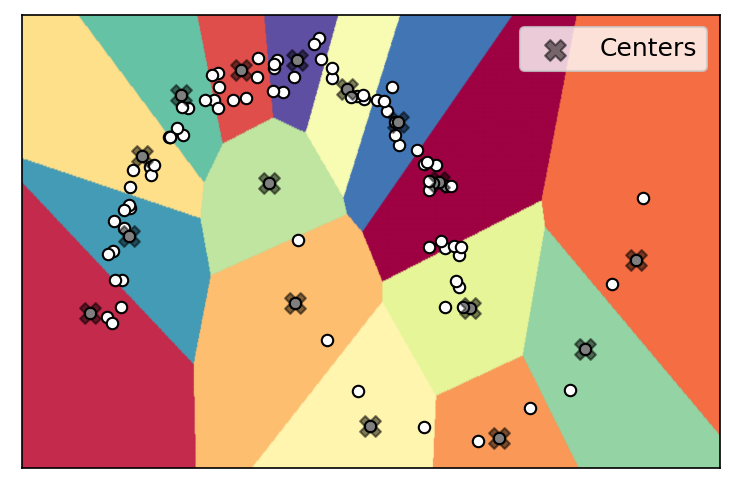}
    \end{minipage}
    \begin{minipage}{.24\textwidth}
    \includegraphics[width=\linewidth]{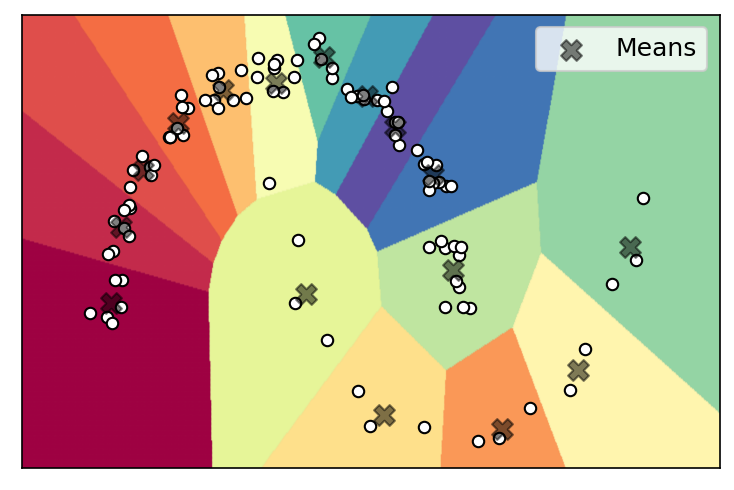}
    \end{minipage}
	\caption{\small From Left to Right, K-means, K-medoids, K-center and cardinality constrained K-means clustering \cite{bradley2000constrained} for the imbalanced two moons summarization task as in Fig.\ref{fig:rconstimbalmoons}. All four of the k-clustering algorithms are biased towards the concave moon with only $\{6,6,6,5\}$ centroids being selected from the convex moon by each method respectively.
% 	See Sec.~\ref{discussion} for details.
	}
	\label{fig.tripletimbalmoons}
\end{figure*}

% \begin{figure*}[t]
% 	\centering
%     \begin{minipage}{.3\textwidth}
%     \includegraphics[width=\linewidth]{100_2moons_kmeans.png}
%     \end{minipage}
%     \begin{minipage}{.3\textwidth}
%     \includegraphics[width=\linewidth]{100_2moons_kmedoid.png}
%     \end{minipage}
%     \begin{minipage}{.3\textwidth}
%     \includegraphics[width=\linewidth]{100_2moons_kcenter.png}
%     \end{minipage}
% 	\caption{\small From Left to Right, K-means, K-medoids and K-center clustering for an the imbalanced two moons summarization task as in Fig.\ref{fig:rconstimbalmoons}. All three of the k-clustering algorithms are biased towards the concave moon with only $6$ centroids being selected from the convex moon.}
% 	\label{fig.tripletimbalmoons}
% \end{figure*}

\textbf{In this paper} we propose radius constrained clustering as a method to introduce robustness to K-means without requiring any domain specific knowledge. That is, the algorithm will produce regions/clusters for which the pairwise distance between samples within that region is upper bounded by a chosen constant.
Our proposed method (Fig.~\ref{fig:rconstimbalmoons}) generates uniformly spaced centroids on the data manifold, while being robust towards sampling inconsistencies. This offers great advantages, e.g., when using K-means to obtain points from a  manifold robust to the distribution of samples. We compare our proposed method with standard k-clustering methods and robust methods such as cardinality constrained clustering \cite{bradley2000constrained} and t-distribution K-means clustering \cite{li2021tk}.
Our contributions in this paper are summarized below:
\begin{itemize}
	\item We present the first k-means algorithm with a hard radius constraint that is tractable. We use a convex relaxation of radius constrained k-means, and pose it as a mixed integer (MI) semi-definite program (SDP). We solve it via a linear SDP relaxation and subsequent rounding.
	\item We present empirical evidence on the efficacy of radius constraints on summarization of data, especially to be robust towards sampling biases.
\end{itemize}
The rest of the paper is organized as follows: in Section \ref{sec.formulation}, we present the radius constraint K-means that we propose, starting from the definition of K-means and moving towards a Mixed Integer Semi-Definite Program (MISDP) formulation of our method. In Section \ref{discussion} we discuss qualitative results comparing with different methods, and, in Section \ref{conclusion}, we discuss future directions.

\section{Background: K-Means}
\label{sec.formulation}

We denote by $\Gamma=\{x_l\}_{l=1}^N$ the set of $N$ data points in $\mathbb{R}^m$. K-means proposes a centroid based clustering, i.e., partition of $\Gamma$ into $k$ disjoint groups found by minimizing
\begin{equation} \label{eq1}
    \min_{\{\Gamma_k\}_{k=1}^K} 
    % \min_{\{\gamma_k\}_{k=1}^K}
    \sum_{k=1}^K \sum_{l\in\Gamma_k} \|x_l-\gamma_k\|^2,
\end{equation}
where $\Gamma_k \cap \Gamma_{k'} = \{\},\forall k \not = k', \cup_{k=1}^{K}\Gamma_k = \Gamma$ and $\gamma_k$ is the centroid of cluster $k$. Let, $\mathbf{1}_{\Gamma_k}$ be the indicator function of the $k$-th cluster, Eq. \ref{eq1} becomes
\begin{align} 
    \sum_{k=1}^K \sum_{l\in\Gamma_k} \|x_l-\gamma_k\|^2 & = \frac{1}{2} \sum_{k=1}^K \frac{1}{|\Gamma_k|}\sum_{l\in \Gamma_k, s \in \Gamma_k} \|x_l-x_s\|^2  \nonumber\\
    &= \frac{1}{2} \sum_{k=1}^K \frac{1}{|\Gamma_k|} \langle \mathbf{1}_{\Gamma_k}\mathbf{1}_{\Gamma_k}^T,\vect{D} \rangle\label{eq2}
\end{align}
where, $\langle \cdot, \cdot \rangle$ is the matrix inner product, and $\vect{D} \in \mathbb{R}^{N \times N}$ is the squared pairwise distance matrix with each element $d_{ij} = \|x_i-x_j\|^2$. 
The first equality in Eq. \ref{eq2} comes from the equality relationship introduced in \cite{zha2001spectral}, relating the sum of pairwise distances with the sum of radial distance for any partition. 
The second equality is a simple matrix reformulation of the inner sum operation. 
Therefore, we can rewrite the k-means problem as,
\begin{equation}
        \min_{\{\Gamma_k\}_{k=1}^K} \frac{1}{2} \sum_{k=1}^K \frac{1}{|\Gamma_k|} \langle  \mathbf{1}_{\Gamma_k}\mathbf{1}_{\Gamma_k}^T,\vect{D} \rangle,
\label{eq3}
\end{equation}
with $\cup_{k=1}^K \Gamma_k = \Gamma$ and $\Gamma_k \cap \Gamma_{k'} = \emptyset$ for $k \neq k'$,
which is an NP hard problem \cite{aloise2009np}. Notice that in the above formulation, there are no explicit constraints on the number of samples per cluster, the intra-cluster radius or the weighting of different samples, e.g., to account for outliers. We propose to take one step into that direction by providing a hard constraint on the intra-cluster radius.

\section{Radius constrained K-Means: No More Than 6ft Apart}
% We define \textit{the radius of as the maximal distance between the center of the largest circle inscribed in a partition maximal distance between the centroid of a partition and any sample residing in it.}
Previously, \cite{zhong2006diameter} have provided formulations for soft radius constraints in online k-means clustering, where the constraint is introduced as an additional term in the optimization objective. We provide a formulation for hard radius constraints $r$, where $r$ is fixed for every cluster. 
% However, specific radius constraint for each cluster can also be introduced without changing the class of the optimization problem.
% For a partitioning $\Gamma = \cup_{k=1}^K \Gamma_k$, we define radius constraints as
% \begin{align}
%         \max \{\|x_l-\gamma_k\|^2  |  l \in \Gamma_k  \} &\leq r^2 & \text{ for } k=1,2,3...,K 
% \end{align}
% where, the radius constraints $r$ is fixed for every cluster, however we will only consider the same value for each cluster in our study. 
Since for any partition with a fixed radius, the maximal distance between two samples can be at most the diameter, we can write the maximal radius constraint as
\begin{equation}
\label{eq6}
\max \{\|x_l-x_s\|^2  |  l,s \in \Gamma_k  \} \leq 4r^2 \; \text{ for } k=1,2,3...,K. 
\end{equation}
The k-means objective in Eq. \ref{eq3} can therefore be rewritten with the maximal radius constraint as,
\begin{align}
    \label{kmeans_w_const}
     \min_{\{\Gamma_k\}_{k=1}^K} & \frac{1}{2} \sum_{k=1}^K \frac{1}{|\Gamma_k|} \langle  \mathbf{1}_{\Gamma_k}\mathbf{1}_{\Gamma_k}^T,\vect{D} \rangle \\ 
        % \text{where} \hspace{1em} Z = \sum_{k=1}^K \frac{1}{|\Gamma_k|} \mathbf{1}_{\Gamma_k}\mathbf{1}_{\Gamma_k}^T \\ 
       \text{s.t} \ &  \cup_{k=1}^K \Gamma_k = \Gamma  ,\;\Gamma_k \cap \Gamma_{k'} = \emptyset \text{ for } k \neq k'  \label{cons2} \\
       & d_{ij} \leq 4r^2 \hspace{.5em} \forall  i,j \in \Gamma_k \text{ for } k=1,2,3...,K. \label{cons3} 
\end{align}
Note that by setting $r^2=\infty$ one recovers the standard K-means form. Before going into the optimization method and empirical validations, we recall that our goal is to leverage the explicit constraint on $d_{i,j}$ to ensure that some regions can not cover samples that are too far apart in the space. 

\subsection{MILP formulation of Radius Constrained K-means}
We start the Mixed Integer Linear Program (MILP) formulation of Eq. \ref{kmeans_w_const} by introducing $NK$ binary variables $\pi_i^k \in \{0,1\}$, where $\pi_i^k=0$ if $x_i \notin \Gamma_k$ and $\pi_i^K=1$ if $x_i \in \Gamma_k$. Therefore, the objective becomes
\begin{align}
    \label{milp_w_const}
     \min_{\forall \pi_i^k} \hspace{1em} & \frac{1}{2} \sum_{k=1}^K \frac{1}{n_k} \sum_{i,j=1}^N d_{ij}\pi_i^k\pi_j^k \\ 
        % \text{where} \hspace{1em} Z = \sum_{k=1}^K \frac{1}{|\Gamma_k|} \mathbf{1}_{\Gamma_k}\mathbf{1}_{\Gamma_k}^T \\ 
       \text{s.t} \hspace{1em} &  \pi_i^k \in \{0,1\}, \hspace{1em} n_k \in \mathbb{Z}, 1 \leq n_k \leq N, \\
       & \sum_{i=1}^N \pi_i^k = n_k, \sum_{k=1}^K n_k = N,  \;\;\sum_{k=1}^K \pi_i^k = 1, \label{cons_milp2} \\
       & d_{ij} \pi_i^k \pi_j^k \leq 4 r^2, \forall i,j,\;\forall k, \label{cons_milp3}
\end{align}
where, $n_k$ are integer variables between $[1,N]$ and $n_k = |\Gamma_k|$ at optimality. It can be easily verified that the constraints \ref{cons_milp2} and \ref{cons_milp3} are equivalent to constraints  \ref{cons2} and \ref{cons3}. 
The feasible set of the original k-means formulation in Eq. \ref{eq1} is also a feasible set of the MILP formulation. 
Our formulation is closely related to the cardinality constrained k-means formulation in \cite{rujeerapaiboon2019size}. 
In our optimization model, we introduce a constraint on the squared pairwise distance inside each partition, while keeping its cardinality as an integer variable; whereas \cite{rujeerapaiboon2019size} allows specifying cardinality constraints for each partition. Note that our proposed model can also allow using different radius constraints $r_k$ in Eq.~\ref{cons_milp3} for different partitions without changing the model class. We avoid that for the sake of simplicity of our formulation. Another thing to note is that the partition radius upper bound $r$ also upper bounds the k-radius, i.e., the maximal distance between any sample and its centroid, by $2r$.

\begin{table}[t]
\centering
\caption{Comparison of partition radius, k-radius and number of k in the convex moon ($\Omega$) for $100$ random seeds. Note that our proposed constrained optimization model ($r=.189$) finds the optimal solution for given constraints.} 
\vspace{0.5em}
\resizebox{\columnwidth}{!}{%
\begin{tabular}{@{}lrcl@{}}
\toprule
Method & \multicolumn{1}{c}{Max Partition Radius} & \multicolumn{1}{c}{Max k-Radius} & k $\in \Omega$ \\ \midrule
K-means & $.228 (\pm .03)$ & $.258 (\pm .04)$ & $5.77 (\pm.42)$ \\
K-medoids & $.268 (\pm .05)$ & $.379 (\pm .07)$ & $5.56 (\pm .68)$ \\
K-center & $.224 (\pm .02)$ & $.289 (\pm .01)$ & $7.52 (\pm .57)$ \\
card. K-means \cite{bradley2000constrained} & $.222 (\pm .03)$ & $.25 (\pm .03)$ & $5.83 (\pm 0.43)$ \\
tk-means \cite{li2021tk} & $.28 (\pm .06)$ & $.309 (\pm .07)$ & $4.92 (\pm 0.63)$ \\
\textbf{Ours} & \textbf{0.181} & \textbf{.207} & \textbf{8} \\ \bottomrule
\end{tabular}
}\vspace{-0.5cm}
\end{table}

% The MILP formulation is intractable and NP hard. \textcolor{magenta}{Citations why?}
% Not proven NP hardness
% We now move towards a convex relaxation of the MILP, namely as a semi-definite program.

\subsection{Convex relaxation of the MILP formulation}
We start the convex formulation by replacing the binary variables $\pi_i^k$ with binary vector $\mathbf{b^k}=\{b_i^k\}_{i=1}^N$,  $b_i^k \in \{-1,1\}$ where $b_i^k$ is 1 if $x_i \in \Gamma_k$ and -1 otherwise. 
This implies $b_i^k = 2\pi_i^k-1$. 
The MILP objective function can be written in terms of $\vect{b^k}$ as:
\begin{align}
    \small
    \label{sdp_objective}
    & \frac{1}{2} \sum_{k=1}^K \frac{1}{n_k} \sum_{i,j=1}^N d_{ij}\pi_i^k\pi_j^k \\
    = \frac{1}{2} \langle \vect{D}, \sum_{k=1}^K \frac{1}{4n_k} & (\vect{M^k} + \mathbf{1}\mathbf{1}^T + \vect{b^k}\vect{1}^T + \vect{1}(\vect{b^k})^T)        \rangle 
\end{align}
% \textcolor{magenta}{Tasos: it is not clear why we have $\approx$ here.}
where $\vect{M^k} \in \mathbb{R}^{N \times N}$. For equality in the feasible set, we need $\vect{M^k} = \vect{b^k}(\vect{b^k})^T$ which yields each element in the right term of the inner product as $ \frac{1}{4n_k} \sum_{k=1}^K b_i^k b_j^k + 1 + b_i^k + b_j^k$. 
% \textcolor{magenta}{Tasos: it might make sense to first re-write the combinatorial problem wrt $\vect{b^k}$, before you relax it to convex.}
The SDP relaxation of the problem therefore is,
\begin{align}
    \label{sdp_w/o_rtl}
     \min_{\forall \vect{b^k},\vect{M^k}} \hspace{1em} & \frac{1}{8} \langle \vect{D}, \sum_{k=1}^K \frac{1}{n_k} (\vect{M^k} + \mathbf{1}\mathbf{1}^T + \vect{b^k}\vect{1}^T + \vect{1}(\vect{b^k})^T)        \rangle \\
       \text{s.t} \hspace{1em} &  -1 \leq b_i^k \leq 1, 1 \leq n_k \leq N,\vect{M^k} \succeq \vect{b^k}(\vect{b^k})^T, \nonumber\\
       & \text{diag}(\vect{M^k}) = \vect{1},\vect{1}^T\vect{b^k} = 2n_k-N, \; \sum_{k=1}^K n_k = N, \nonumber\\
       & \sum_{k=1}^K \vect{b^k}\hspace{-0.04cm} =\hspace{-0.04cm} (2-K)\vect{1}, 
       d_{ij}(m_{ij}^k \hspace{-0.04cm} +\hspace{-0.04cm} 1\hspace{-0.04cm} + \hspace{-0.04cm}b_i^k\hspace{-0.04cm} +\hspace{-0.04cm} b_j^k) \hspace{-0.04cm}\leq\hspace{-0.04cm} 16 r^2, \nonumber\\
       \text{for} \hspace{1em} & i,j=1,2,...,N \text{ and } k=1,2,...,K \nonumber 
\end{align}
% where, the only quadratic components in the formulation arise in
where, the semi-definite constraint $\vect{M} \succeq \vect{b}(\vect{b})^T$ can be converted into a linear matrix inequality using Schur's complement \cite{zhang2006schur}.
% We also add the following constraints to lower and upper bound $\vect{M^k}$,
% \begin{align}
%     \vect{M^k} + \mathbf{1}\mathbf{1}^T + \vect{b^k}\vect{1}^T + \vect{1}(\vect{b^k})^T  & \geq 0 \nonumber \\
%     \vect{M^k} + \mathbf{1}\mathbf{1}^T - \vect{b^k}\vect{1}^T - \vect{1}(\vect{b^k})^T & \geq 0 \label{eq.rtl}\\
%     \vect{M^k} - \mathbf{1}\mathbf{1}^T + \vect{b^k}\vect{1}^T - \vect{1}(\vect{b^k})^T & \leq 0 \nonumber \\
%     \vect{M^k} - \mathbf{1}\mathbf{1}^T - \vect{b^k}\vect{1}^T + \vect{1}(\vect{b^k})^T & \leq 0. \nonumber
% \end{align}
\begin{figure*}[t!]
	\centering
	\includegraphics[width=1\textwidth]{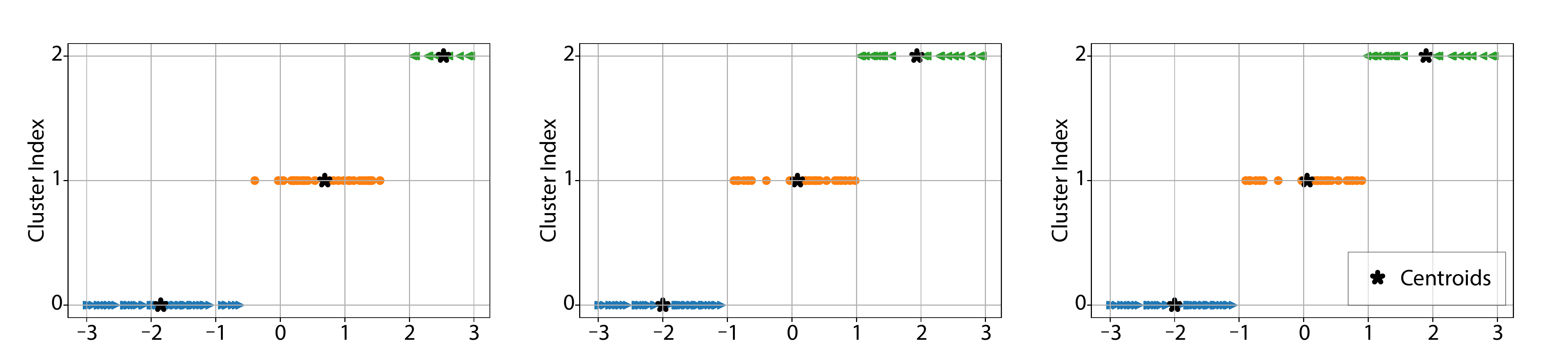}
	\vspace{-0.7cm}
	\caption{\small From Left to Right, K-means, K-means with radius constraint$=1$ and K-means with maximal cardinality constraint$=55$.}
	\label{fig.1Dk3}
\end{figure*}

% \vspace{-12em}
\begin{algorithm}[t!]
	\SetAlgoLined
	\KwIn{$N; \vect{x} = \{x_i\}_{i=1}^N; K; r; \vect{D} \in \mathbb{R}^{N \times N}$}
	\KwOut{binary assignments $\vect{\Pi} \in \{0,1\}^{N \times K}$}
	Step 1: Solve continuous relaxation of MISDP using interior point method $\rightarrow \vect{b^k}$  $\forall k $ \;	Step 2: Let, $\vect{\Pi} \in \{0,1\}^{N \times K}$ binary partition assignment variables. Solve,
	\begin{align}
		\label{eq.quad1}
		\max_{\vect{\Pi}} \hspace{1em} & \sum_{k=1}^{K} \sum_{i=1}^{N} \pi_i^k b_i^k \;\; \text{ s.t } \hspace{0.5em}  1 \leq \sum_{i=1}^{N} \pi_i^k \leq N, \\
	 \text{and } & \sum_{k=1}^{K} \pi_i^k = 1 ,\;d_{ij} \pi_i^k \pi_j^k \leq 4 r^2 \hspace{1em}\forall i,j,k \nonumber
	\end{align}

	Step 3: Get partitions $\Gamma_k$ and centroids $\gamma_k \hspace{1em} \forall k$
% 	from $\vect{\Pi}$ and $\vect{x}$\;

	Step 4: Solve for $\vect{\Pi} \in \{0,1\}^{N \times K}$
	\begin{align}
		\min_{\vect{\Pi}}  \hspace{-0.05cm}& \sum_{k=1}^{K} \sum_{i=1}^{N} \hspace{-0.05cm}\pi_i^k  \|x_i\hspace{-0.05cm}-\hspace{-0.05cm}\gamma_k\|^2\text{ s.t }   1 \hspace{-0.05cm}\leq\hspace{-0.05cm} \sum_{i=1}^{N} \pi_i^k\hspace{-0.05cm} \leq\hspace{-0.05cm} N
		,\label{eq.quad2}\\
		 &\text{ and }\sum_{k=1}^{K} \pi_i^k = 1, d_{ij} \pi_i^k \pi_j^k \leq 4 r^2,\forall i,j,k\nonumber
	\end{align}
	\caption{Maximal radius constrained k-means}
	\label{alg.kmeans_rconst}
	\vspace{-0.3cm}
\end{algorithm}
% We also add constraints to lower and upper bound $\vect{M^k}$ which comes from reformulation-linearization technique \cite{anstreicher2009semidefinite}.
The objective in the current formulation is a linear fractional function which can be turned into a linear objective using Charnes-Cooper transformation \cite{charnes1962programming}. 
Specifically, since the denominator in Eq. \ref{sdp_objective} is strictly positive as $n_k \geq 1$, the objective can be expressed as a perspective function \cite{boyd2004convex}. 
Without the continuous relaxation of $\vect{b^k}$, the problem can be formally stated as a Mixed Integer Semi-definite program (MISDP) \cite{gally2018misdp}. 
% \textcolor{magenta}{Tasos: is this a well-known class? Are there other problems that use MISDP? Citations?}
% In the following section, we discuss the rounding algorithm we propose to find a solution for Eq. \ref{milp_w_const}.

% \begin{figure*}[b]
% 	\centering
% 	\includegraphics[width=\textwidth]{1Dk3_.pdf}
% 	\caption{\small From Left to Right, K-means with radius constraint of $r=1.2$,$r=1.1$ and $r=1$. As the radius constraint is tightened, the maximal radius of cluster $0$ decreases while the radius for cluster $2$ starts increasing to adhere to the coverage constraint. At $r=1$ the clusters become uniformly sized.}
% 	\label{r_sweep}
% \end{figure*}

\subsection{Rounding Algorithm}

% \textcolor{magenta}{Tasos: definitely, there is "meat" here to describe better. E.g., your algorithm seems like a meta-algorith, where one step is solving MISDP that you have been discussing so far. Describing the rest steps better + describe why all 5 steps together make sense is something to be added. I leave this section intact.}

We define our rounding algorithm as a two-step linear assignment problem with quadratic constraints (Alg. \ref{alg.kmeans_rconst}). The first step in the algorithm is to find binary partition variables for each sample, we define it as $\vect{\Pi} \in \{0,1\}^{N \times K}$ where each element $\pi_i^k$ is $1$ if $x_i$ is assigned to cluster $k$. We solve two linear assignment objectives, one in which we maximize the inner product sum of the SDP solution and $\pi_i^k$ with quadratic constraints adhering to the maximal radius constraint (Eq. \ref{eq.quad1}). In the second step, we minimize the intra-cluster distance for the assignment variables (Eq. \ref{eq.quad2}). 
% The algorithm details are presented in Alg. \ref{alg.kmeans_rconst}. Note that Step 2-4 in our algorithm is similar to one lloyd step in k-means. 
% \vspace{-2em}
\section{Experiments}
\label{discussion}

\textbf{2D experiments.} Experimental results presented in Fig.~\ref{fig:rconstimbalmoons} and Fig.~\ref{fig.tripletimbalmoons} portray the efficacy of radius constrained k-means for imbalanced data summarization. We compare with cardinality constrained k-means \cite{bradley2000constrained} as an alternative constrained k-means method. We also compare with tk-means \cite{li2021tk} which uses long tail assumptions for robustness. For comparison, we sweep the cardinality upper and lower bounds of \cite{bradley2000constrained} till infeasible to find the best balance. We see that both increasing the lower or decreasing the upper bounds from respectively $0$ and $N=100$ harms balanced centroid generation; increasing the lower bound makes it easier to achieve the lower bound for the convex moon, while decreasing the upper bound requires more centroids to cover the concave moon. For all experiments we choose $N/K$ to be small since otherwise, k-means based clustering might return centroids off the data manifold, therefore yielding bad sketches/summaries. An added benefit radius constraints provide is a feasibility certificate- tighter radius bounds resulting in empty solution sets can be used to infer how to increase $K$ to be able to cover all the samples. For the imbalanced two moons experiments, for a radius constraint of $r=.189$, we have seen that at least $16$ centroids were required to be able to cover the whole manifold. For different methods, the partition radius and k-radius is presented in Table 1.

\textbf{1D experiments.} Let, we have $N=102$ samples from three uniform distributions $\textit{U(-3,-1)}$, $\textit{U(-1,1)}$ and $\textit{U(1,3)}$ with $51, 26, 25$ samples in each respectively. We draw comparisons between standard k-means, cardinality constrained k-means and radius constrained k-means in a $k=3$ summarization task. 
% Suppose, this is a simulation where, each of the uniform distributions represent a race and they are sampled inconsistently.
Fig.~\ref{fig.1Dk3} shows clustering performance for such a case; without any constraints, k-means will create an inconsistent partition, e.g., resulting in the mixing of different attributes represented by each random variable. This will yield centroids which are not proper summaries of the dataset. Whereas with radius constraint of $1$ and cardinality constraint of $55$ adhere to the correct partitioning. In such settings, cardinality constrained k-means require re-tuning the constraint when the dataset is resampled, whereas ours is robust.

\textbf{Implementation.} We use \texttt{MOSEK} to solve Step 1 in Alg.~\ref{alg.kmeans_rconst} and \texttt{Gurobi} to solve Steps 2 and 4. In our experiments, we did not require tuning of solver parameters. 

% Please add the following required packages to your document preamble:

% \textbf{Maximal and Minimal Constraints.} Even though our formulation employs only a maximal constraint on the partitions, for a fixed $k$ constraining the maximal radius of a partition would implicitly require another partition to maintain a minimal radius such that all the data samples are covered by the partitions. This dual relationship between the maximal and minimal constraints can be employed in different applications such as local differential privacy as well as privacy guarantees similar to k-anonymity. Fig. \ref{r_sweep} portrays how tuning the maximal radius constraint changes the minimal radius requirement of the smallest cluster.

\section{Conclusion}
\label{conclusion}

We propose the first maximal radius constrained K-means as an MISDP optimization objective. Upon comparison with multiple k-clustering methods, we see that our method is more robust towards sampling bias/ data imbalance. The main limitation of our radius constrained k-means formulation is that both the order of variables are $\mathcal{O}(k.N^2)$ which is impractical for very large datasets. From preliminary experiments, we see that replacing the SDP problem with a k-center problem in Step 1 of Alg.~\ref{alg.kmeans_rconst} has minimal effects on the centroid selection. This can be considered a future direction to improve computational complexity.

\section*{Acknowledgements}
Humayun and Baraniuk were supported by NSF grants CCF1911094, IIS-1838177, and IIS-1730574; ONR grants N00014- 18-12571, N00014-20-1-2534, and MURI N00014-20-1-2787; AFOSR grant FA9550-22-1-0060; and a Vannevar Bush Faculty Fellowship, ONR grant N00014-18-1-2047.

\bibliographystyle{IEEEbib}
\bibliography{refs}

\end{document}